# Semantic-ontological combination of Business Rules and Business Processes in IT Service Management

Alexander Sellner[1], Christopher Schwarz[1], Erwin Zinser[1]

[1]FH JOANNEUM University of Applied Sciences, Department of Information Management
Alte Poststrasse 147, 8020 Graz, Austria
`{Alexander.Sellner|Christopher.Schwarz|Erwin.Zinser}`
`@fh-joanneum.at`

**Abstract.** IT Service Management deals with managing a broad range of items related to complex system environments. As there is both, a close connection to business interests and IT infrastructure, the application of semantic expressions which are seamlessly integrated within applications for managing ITSM environments, can help to improve transparency and profitability. This paper focuses on the challenges regarding the integration of semantics and ontologies within ITSM environments. It will describe the paradigm of relationships and inheritance within complex service trees and will present an approach of ontologically expressing them. Furthermore, the application of SBVR-based rules as executable SQL triggers will be discussed. Finally, the broad range of topics for further research, derived from the findings, will be presented.

## 1   Introduction

IT Service Management (ITSM) can be seen as a large and complex environment for business processes and rules with a certain potential for automation. On the one hand, ITSM puts a strong focus on providing tools for managing business topics such as outsourcing costs, licensing fees and negotiated prices within Service Level Agreements (SLAs), on the other hand the goal is to provide clearly defined processes for managing IT resources, which are, for instance defined in the ITIL framework [1].

Having a look at ITSM from large IT service providers' perspectives, there is an tremendous amount of so-called configuration items which are combined together to services, which are ultimately sold to clients. Due to changes in system environment or technological development, these services are in a constant state of change, turning the task of outsourced service provision to a rather stable price to a quite difficult challenge.

Most ITSM activities affecting existing relationships between IT service providers and their clients are triggered by so called "requests for changes" (RFCs). RFCs usually add, remove, or change existing services e.g. upsize the RAM of a server or add an SLA to a database application. Almost every ITSM item will have a certain dependency to another ITSM item and, depending on the complexity of the services, these relationships will lead to the formation of a so-called service tree. The service tree can be seen as one or more graphs, as there is no single parent node and items, such as SLAs can exist multiple times.

A well-suited approach for making such graphs human-readable is to make use of ontologies using semantic expressions [2]. By applying ontologies, it becomes possible to create graphical representations of the complex service trees, making it possible to discover all dependencies and to keep them well-managed. A further advantage becomes obvious, having a look at natural-language based semantic expressions used by ontologies. Besides adding further advances to the comprehensibility and integrity of service trees, this also creates the possibility of performing commonly understandable modifications of service relationships and therefore can help to create a mutual understanding between ITSM service providers and their customers [3].

Because of the strong business connection and great involvement of rules within SLAs, the idea of establishing a rule repository for service level definitions which are based on natural language, seems obvious. Over the last years, several standards for such definitions have been created such as RuleSpeak[1], R2ML[2] and Semantics of Business Vocabulary and Business Rules (SBVR)[3]. Most recently, it seems that there is a strong support for the SBVR standard both within the academic community as well as the industry.

This paper focuses on the challenge of displaying complex graphs of service relations to human-readable ontologies, based on semantic models. Furthermore the paper will discuss on the special topic of enriching these complex service trees with SLAs assembled by SBVR-based business rules. Another topic being discussed will be the establishment of SBVR-based business rules using DBMS triggers for execution [4].

The paper is structured as follows: Section 2 will have a closer look at service trees, discuss the paradigm of service inheritance and present ways of ontologically displaying existing service structures. Section 3 will explain the main categories of SLA rules within service trees and present samples for converting SBVR definitions to database-executable statements. Section 4 will discuss topics for further research within the presented areas.

---

[1] www.rulespeak.com/
[2] http://rewerse.net/
[3] http://www.omg.org/spec/SBVR/1.0/

## 2 Establishing ontological views of service trees based on existing semantic definitions

As already briefly discussed in the previous section, the structure of service trees in real-world business environments can be quite complex and is therefore hard to handle and maintain by people carrying out various tasks within ITSM processes. The following example will provide closer insights into the process of establishing RFCs as well as the related legacy data and the necessary procedures to involve ontological views which allow establishing a well-structured perspective on the configuration items' relationships and corresponding inherited attributes.

Within the given case, which is based on a project carried out for a large ITSM provider, the following types of configuration items (CIs) can be identified:

- Service: A service is a collection of ITSM assets with a business-relevant impact. This means that they are the elementary constructs within contracts between service providers and customers. Having a look at the function of services within the service tree, they can also be used as logical containers for putting together various "low-level" services to more sophisticated "high-level" services. Therefore, a low level service can for instance be a domain name resolution service and a high level service could be a billing application. As a matter of fact, services can exist multiple times within the service tree.
- Host: Hosts are the fundamental basis for establishing services. As a result, a service can be put together by one or more hosts. Because of a strong advance in virtualization technology over the recent years, hosts don't necessarily need to be actual hardware but multiple hosts could also depend on a host-service relationship providing the actual hardware for virtualization as well as on an additional host-service relationship for data storage. The same host can be used by multiple services and therefore can appear multiple times within the service tree.
- Service Level Agreement: SLAs include legal implications regarding the quality of services. For instance, the loss of an email service for a certain time could have a significant business impact for organizations. As a result, SLAs also handle penalty fees which need to be paid under certain circumstances. This can, for instance, be as soon as a service becomes unavailable (first failure fines and concurrent failure fines) or if a certain value for the availability of a service over a certain period is not met (availability percentage per day, month or year). SLAs follow the principle of inheritance and are therefore valid for all services or hosts which are hierarchically subordinate within the service tree. In addition, SLAs can appear multiple times within the service tree and they need to be prioritized in order to prevent discrepancies by allowing only one valid SLA per host or service.
- Maintenance Contract (MTC): MTCs contain contractually defined regulations regarding repair and renewal tasks on hardware. This can for instance be the replacement of a defect part or the regular upgrade to newer hardware parts. MTCs can be closed for services or hosts have the same characteristics as SLAs regarding inheritance. In contrast to SLAs, there can be multiple MTCs valid at the same time for one service or host which leads to accumulated maintenance liabilities.

Besides these CIs, there can be a wide range of additional CIs involved in real-world ITSM processes [5]. The above-mentioned items will serve to establish a simplified demonstration of a sample ITSM service tree which already contains a great variety of characteristics worth to undergo closer investigation. The sample service construct is demonstrated in Figure 1.

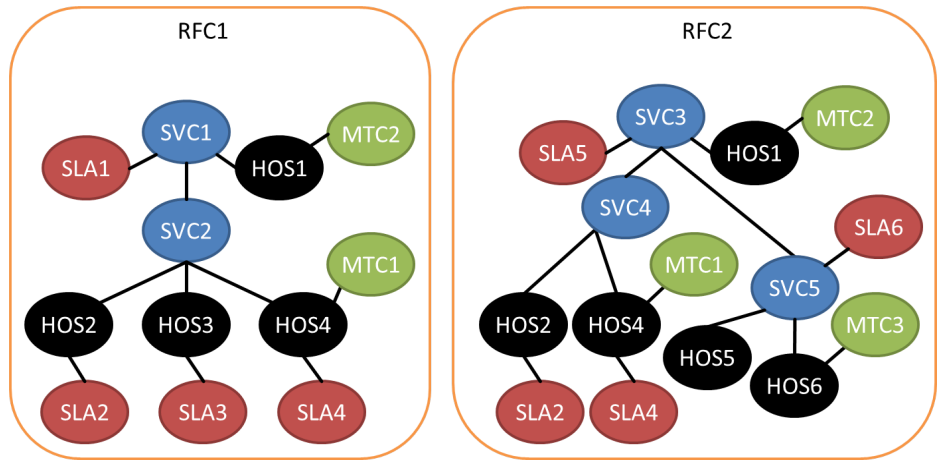

**Figure 1.** Sample ITSM service construct

The sample demonstrates two RFCs with their respective hierarchically subordinate services, hosts, SLAs, and MTCs. RFC1 might, for instance, be a database service hosted on a virtual machine (HOS1) with HOS2, HOS3 and HOS4 being redundant database servers, each having different SLAs. RFC2 could be a Web-based application consisting of SVC4 as database backend and SVC5 as Web servers, whereas SVC4 consists of two machines (HOS2 and HOS4) already being used by RFC1.

Considering the characteristic of inheritance for SLAs and MTCs within service trees, the example clearly illustrates why more complex service trees are almost impossible to manage and handle from a business perspective.

Having a look at the SLAs in RFC1, it clearly needs to be determined whether SLA1 or one of the respective SLAs for HOS2, HOS3 and HOS4 is valid. This is a prime example where priorities need to be assigned to SLAs in order to ensure legal correctness and optimized business benefit. Such a prioritization even needs to be put in place across boundaries of RFCs. For instance within RFC2 it also needs to be determined whether SLA4 or SLA2 and SLA5 respectively are valid for HOS2 and HOS4. Changing priority of SLA2 or SLA4 would again have an impact on the overall situation of RFC1

When focusing closer on the question of how prioritization between two SLAs needs to be done, it becomes clear that sometimes it is quite difficult to say which SLA is preferable to be applied . Assuming that within RFC1, the fines for first failure are higher in SLA1 than in SLA2 and the availability fines are higher in SLA2 than in SLA1, it is impossible to instantly determine the most cost effective solution. It is rather imaginable that the availability of service or hosts is monitored over a certain period leading to statistical forecast regarding its behavior. Only then it would be possible to pick the optimal SLA, taking all types of fines into account.

In contrast, the challenge regarding MTCs is rather on deciding whether individual MTCs should be discharged because of redundancies or existing MTCs should be extended. As a result, the accumulated MTC benefits should be considered for every individual host or service.

## 3  Invoking semantics within ITSM service trees

To overcome the challenge of managing complex relations within ITSM service trees, it seems quite obvious to introduce ontologies and semantic expressions. This makes it possible to display dependencies between configuration items in a human-readable way and supports extending or rearranging the service tree [6]. An example of such an ontological view taken from the prototype of an ITSM application of a large IT service provider is given in Figure 2. The figure displays an RFC which is linked to an SLA, an MTC, a hardware asset and a high level infrastructure service which is then linked to a low level infrastructure service. The figure also reveals the concept of service instances used to express configuration items which appear multiple times within a service tree.

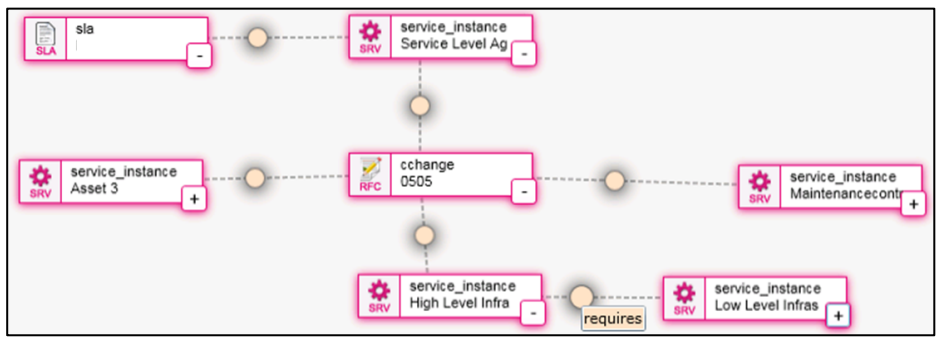

**Figure 2.** Ontological view of ITSM service tree

Integrating semantic expressions, adds a further level of complexity regarding the underlying data structure [7]. This can lead to quite a heavy integration effort regarding database design and operations performed to read, write or update data.

On the other hand, the advantages for integrating semantic expressions are obvious. It becomes possible to create human-understandable illustrations of service trees based on natural language. Additionally, rules can be applied on the service structure based on the existing semantic expressions. Consequently, these rules can contribute to handle the situation regarding SLA and MTC inheritance within the service tree.

## 4   Integrating SBVR-based rules using DBMS triggers for execution

In general, three categories of rules can be identified within ITSM service trees:

1. Rules to prevent adding new SLAs or MTCs due to inconsistencies or disadvantageous business impacts
2. Rules to analyze and improve existing SLA and MTC structure
3. Rules within SLAs themselves

There are various ways following the goal of achieving a loosely coupled execution of these SLA rules within a repository on the specified service tree. A quite feasible approach is to aim for rule execution within Database Management Systems (DBMS) using SQL triggers. This approach leverages characteristics of DBMS to establish as rule engines allowing the execution of Event-Condition-Action (ECA) rules [8].

Recent findings dealing with the conversion of SBVR definitions to SQL statements were taken as basis for developing a prototype making it possible to express the first category of rules for SLAs in structured English and carry out the respective database operations to put them in action [9,10].

Listing 1 gives an example for the conversion of the first category of rules from structured English to an ECA rule which is placed on the DBMS.

---

Structured English statement:

```
T:SLA
T:SVC
T:total fines
F: SLA has total fines
F:SLA is linked to SVC
NR: For an SLA that is linked to an SVC it is obligatory that
the total fines of the new SLA are less than the total fines
of the old SLA.
```

SQL expression:

```
CREATE TRIGGER "NR1" BEFORE UPDATE OF "SLA_id"
  ON "SLA-is_linked_to-SVC"
WHEN NOT
(SELECT "total fines" from "SLA" where id=new.SLA_id)<
(SELECT "total fines" from "SLA" where id=old.SLA_id)
  BEGIN
    SELECT RAISE(ABORT, 'Requirement of NR1 not met');
  END;
```

**Listing 1.** Conversion from structured English to DBMS trigger statement

The terms "SLA" and "SVC" and "total fines" are expressed as lines starting with "T". The fact types "SLA has total fines" and "SLA is linked to SVC" are marked by a preceding "T". Finally, the normative rule is denoted by "NR".

Regarding the definition of fact types, the identification of table attributes takes place using conjugations of the predicate "have" and a table relationship is established by the predicate "being linked to".

By choosing "NR" as markup, the creation of the SQL trigger is either based on a fact type involving a relationship or on a table related to a single term. The identification is based on the phrase being placed in front of the rule. If a fact type is identified as input, the first term used to describe the identity column for the SQL trigger.

Based on the fact type definition for the attribute "total fines", the reverse order "total fines of the new SLA" forms that basis for the SELECT statement (SELECT "total fines" from "SLA" where id=new.SLA_id), whereas the definitions "new" and "old", are explicitly required because of the targeted syntax of the SQL trigger.

Because of the fact that the trigger should not abort when the defined rule is met, the WHEN clause needs to be followed by NOT in order to display an error message in any other cases.

## 5    Conclusions and Future Work

The approach presented, shows the advantages and challenges regarding the application of ontologies and semantic expressions. Findings derived, led to conclusions regarding prerequisites which need to be fulfilled, especially when aiming for a close connection to underlying data structure.

Adding semantic expressions to ITSM service trees can increase presentability and manageability. On the other hand, such an integration also implies a challenge towards the underlying data, because of the paradigm of inherited SLAs and MTCs within the service tree.

The application of SBVR based business rules adds further needs which must be met by database models. In fact, the database design needs to be carried out in accordance with naming conventions established within SQL statements converted from SBVR.

Having a look at the originating statements defined in structured English, a strongly controlled natural language needs to be applied in order to enable a conversion to SQL definitions. The approach presented, focused on SBVR-based rules for preventing negative business impacts in connection with the addition of new SLAs. It needs to be investigated how these statements need to be verbalized for MTCs. It will also be necessary to look at the conversion to stored procedures, which would allow apply rules not only on a data manipulation level.

Further research will also be necessary to mathematically investigate paradigm of inheritance within ITSM service trees. Most likely, this will be closely related to graph theory and provide further insights how service trees can be optimized.

Another area worth performing research on will be investigating the application of upper level ontologies within the specific scenario. This might lead to an enterprise ontology for combining business processes and rules or to an ontology being closely focused on the area of ITSM.